# From "What" to "When" - a Spiking Neural Network Predicting Rare Events and Time to their Occurrence

**Mikhail Kiselev**
Chuvash State University
Cheboksary, Russia
`mkiselev@chuvsu.ru`

## Abstract

The temporal dimension is important for majority of real-world tasks related to prediction of future events. In most cases, it is not sufficient just to predict that some event will happen – it is necessary to evaluate when it will probably happen. Since spiking neural networks (SNN) can naturally and explicitly include time in information processing performed by them, they seem to be an appropriate tool for solution of the event prediction problem. Solution of this problem is important for reinforcement learning (RL) – for prediction of receiving a reward or punishment signal, especially under the quite common condition when these signals are rare. The ability to predict receiving reward in the near or more distant future means the ability to evaluate the current state as more or less close to the target state (labelled by the reward signal). In the present work, we utilize a spiking neural network (SNN) for solution of this problem. In the context of SNNs, events are represented as spikes emitted by network neurons or input nodes. It is assumed that target events are indicated by spikes emitted by a special network input node. Using description of the current state encoded in the form of spikes from the other input nodes, the network should predict approximate time of the next target event. This research paper presents a novel approach to learning the corresponding predictive model by an SNN consisting of leaky integrate-and-fire (LIF) neurons. The proposed method leverages specially designed local synaptic plasticity rules and a novel columnar-layered SNN architecture. Similar to our previous works, this study places a strong emphasis on the hardware-friendliness of the proposed models, ensuring their efficient implementation on modern and future neuroprocessors. The approach proposed was tested on a simple reward prediction task in the context of one of the RL benchmark ATARI games, ping-pong. It was demonstrated that the SNN described in this paper gives superior prediction accuracy in comparison with precise machine learning techniques, such as decision tree algorithms and convolutional neural networks.

***Keywords***: Spiking neural network, spike timing dependent plasticity, dopamine-modulated plasticity, anti-Hebbian plasticity, reinforcement learning, neuromorphic hardware, columnar-layered network architecture

## 1 Introduction

The last two decades, spiking neural networks (SNN) attract growing attention of artificial intelligence researchers as more biologically plausible models than traditional artificial neural networks (ANN). In the process of research, it was discovered that SNN can have other advantages, namely, possibility of extremely low energy consuming hardware implementation and more natural approach to learning and expressing temporal aspects of external events and control commands produced by the network. The latter advantage is explained by the fact that the time dimension explicitly enters information processing performed by SNN – spiking neuron are dynamic systems whose parameters change in time in accordance with certain ODEs. Besides that, the signal propagation time from neuron to neuron is not zero and plays an important role. Therefore, application of SNN to time-related learning tasks seems to be quite natural. The present paper is devoted to one class of these tasks, namely, prediction of time when certain target event will most probably happen. In this problem, SNN receives information about external world in the form of impulses (spikes) from its input nodes. Besides that, one its special input node signals by spike about the event which should be predicted. Using these signals, SNN should learn to predict approximate time to the next target event from the input signal.

Since SNN is a discrete system operating with spikes, it cannot produce the required prediction in the numeric form. Instead, there should be special output neurons in it, such that a spike emitted by a certain output neuron should indicate that the next target event is expected in the time interval in the future corresponding to this neuron. We assume that these time intervals are of the same length *L*, do not intersect and cover all future up to certain depth. Also, it is assumed that two subsequent target events are separated by time period much greater than *NL*, where *N* is the interval count.

Let us rigorously formalize the accuracy criterion for our task. Let us denote the time moment of *i*-th target event as $T_i$. For every time moment *t*, we define time to the next closest $T_i$ as $T(t)$. The network described below should learn to predict the discretized value of the temporal proximity to the next $T_i$:

$P(t) = \max\left(N - \left\lfloor T(t)/L \right\rfloor, 0\right)$. These predictions $P^*(t)$ are made based on activity of *N* network's output neurons using the following iterative formula (assuming that t and $T_i$ are integers, $t \geq 0$, $T_i > 0$ and $P^*(0) = 0$):

$$P^*(t+1) = \begin{cases} 0 \text{ if } \exists_i t = T_i \\ n \text{ if } \exists_i t+1 = T_{ni} \\ 0 \text{ if } \forall_{ni} T_{ni} < t - L \vee T_{ni} > t+1 \\ P^*(t) \end{cases}, \quad (1)$$

where $T_{ni}$ is the moment of *i*-th firing of *n*-th output neuron. We consider the rare event prediction problem in the regression terms with the predicted variable *P(t)* and the predictions *P\*(t)*. Namely, the SNN should learn to maximize the coefficient of determination

$$R^2 = 1 - \frac{D[P^*(t) - P(t)]}{D[P(t)]}. \quad (2)$$

This goal is achieved by an SNN consisting of leaky integrate-and-fire (LIF) neurons with specially designed structure and novel synaptic plasticity rules conforming to the locality principle which declares that value of synaptic weight modification must depend on parameters of pre- and post-synaptic neuron activity only. These features are described in detail in the following section.

We would like to note that our problem lies between two prominent problems attracting much attention of machine learning and neural network researchers – time series forecasting [1 - 3] and anomaly detection [4]. It is located between them but intersects with neither of them. Time series forecasting techniques predict future value of certain variable (discrete or continuous) from the current and recent values of this and, maybe, some other variables. However, we are not interested in knowledge of future values of various parameters characterizing world state, at least, at the present stage of our research. Instead, we would like to know that some rare event (reward) may occur sufficiently soon. On the other hand, similar to our case, anomaly detection algorithms deal with rare events. But usually, these are true anomalies – the events which occur suddenly, without any observable causes – like a sudden hacker attack interrupting normal routine operation of a corporate computer network. The goal of these algorithms is rather to detect them early than to forecast them (that is hardly possible). However, a reward signal in RL while being a rare event is not an anomaly – it is a consequence of previous world states and network actions so that it is quite possible to predict it. Such a specific position of our problem explains the fact that there are almost no works devoted to application of SNNs to its solution – it is located in a "darker", less covered by research location neighboring with highlighted regions of time series prediction and anomaly detection.

One of such SNN applications is described in [5]. It utilizes the NeuCube system [6], which is based on the Liquid State Machine concept (LSM) [7]. The LSM is a large, chaotic, non-plastic SNN transforming a combination of time series data and static (or slowly changing) parameters into a high-dimensional representation in the form of the current firing frequencies of neurons within the LSM. Due to the great number of neurons in the LSM, representations of various spatiotemporal patterns in the form of LSM neuronal activity are linearly separable with high probability. Consequently, classification problems related to such representations can be efficiently solved using simple linear classifiers. As described in [5], several instances of NeuCube's application to rare event forecasting have been demonstrated. One specific example, the prediction of strokes, is examined in greater depth in [8]. While the LSM-based approach has demonstrated success across a wide range of tasks, it has a notable drawback in that it requires a large, computationally intensive LSM to achieve efficiency. In contrast, as we will show below, our network solving rare event prediction problems can be very small.

As it will be discussed in the next section, in order to provide our SNN with the ability to learn to predict rare events, we employ a combination of two synaptic plasticity models—commonly referred to, albeit in a very approximate sense, as Hebbian plasticity and dopamine plasticity. The well-known example of Hebbian plasticity is the STDP plasticity model (Spike Timing Dependent Plasticity) [9]. Dopamine plasticity is typically associated with reward-related effects. The collective terminology for these merged plasticity models is R-STDP (Reward Spike Timing Dependent Plasticity). Numerous works have explored and proposed various R-STDP models [10 - 13], with some of them having already undergone testing in real-world applications [14]. Presently, there is no universally accepted consensus on how to best combine these two types of synaptic plasticity, resulting in a wide array of proposed models. Usually, unlike our approach where reward takes the form of a spike signal, these models use the reward representation a global real-valued variable. To the best of my knowledge, in no prior work (except our previous work [15]) a similar form of R-STDP has been explored. Furthermore, our specific objective was to design the plasticity model in such a way that would facilitate efficient implementation on contemporary and forthcoming neuroprocessors. According to work [16], there is a clear trend in the emergence and development of software and hardware systems that allow not only the inference of convolutional neural networks converted into SNN form, but also the use of arbitrary models of spiking neurons and plasticity rules, opening up the possibility of continuous learning.

In the subsequent sections, we will describe the SNN architecture and the synaptic plasticity model, which combines Hebbian (effectively, anti-Hebbian) and dopamine plasticity. Following this, we will consider the application of this SNN to the problem of reward prediction in reinforcement learning (RL), using as an example the task of reward prediction in the "ping-pong" RL task. Subsequently, we will assess the advantages and limitations of our approach and outline future research plans in this direction.

## 2 Materials and Methods

As it was said in Introduction, our SNN should have two kinds of input nodes – sensory input describing the current state of the external world and the single special input node, spikes from which indicate target events. The network consists of $N$ sections (we will call them columns) which have the same structure (Fig. 1). Each section includes a neuron (it is denoted as SECREW on Fig. 1) which serves as an SNN output neuron mentioned in Introduction. Thus, the network has $N$ output neurons indicating $N$ levels of proximity to the next target event. The network should learn to react to the external stimulation by spikes from SECREW neurons which would maximize criterion (2).

Our neuron model is based on a very simple neuron model called leaky integrate-and-fire (LIF) neuron with current-based delta-synapses. Every time the neuron obtains a spike via its synapse with the synaptic weight *w*, its membrane potential *u* instantaneously changes by the value *w*. Without incoming spikes, *u* decays exponentially to 0 with the time constant $\tau$. If some input spike elevates *u* to the value 1 or greater, then the neurons fires and *u* is reset to 0.

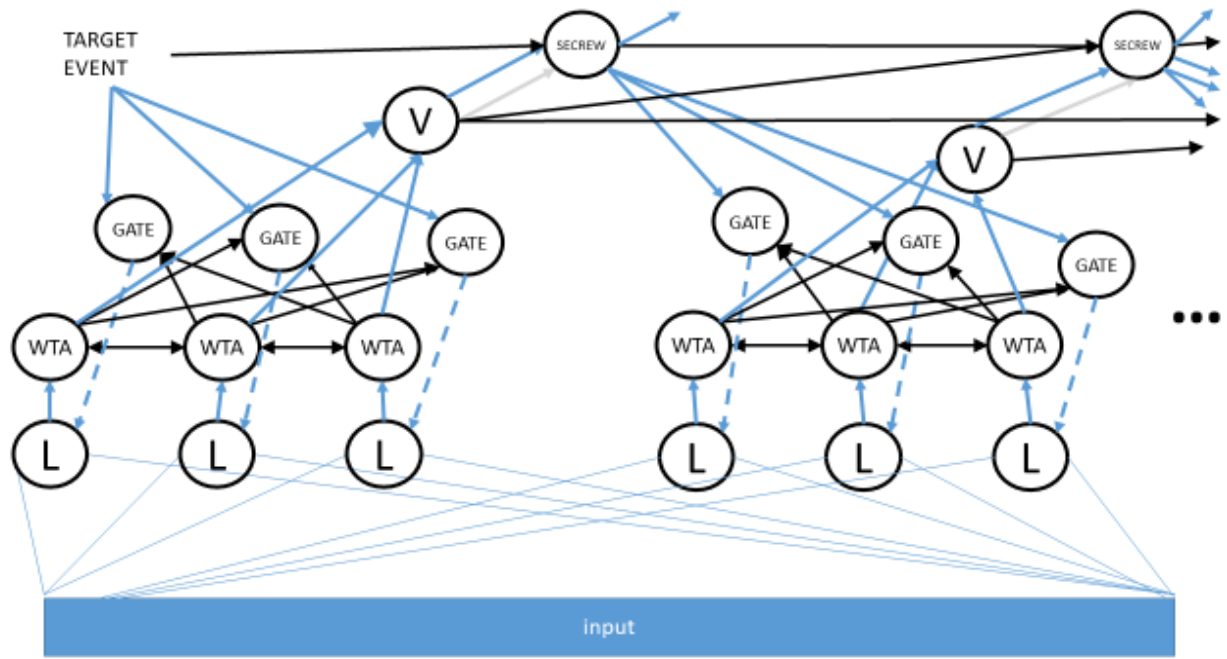

Fig. 1. Structure of the SNN learning to predict rare events. The detailed explanations are in text.

Our neuron model has only one important addition to the standard LIF model – our neuron has special synapses called *blocking*. Whenever a neuron obtains a spike via its blocking synapse it becomes *inactive* for the time period equal to the weight of this blocking synapse. In inactive state, incoming spikes do not affect the membrane potential and, therefore, neuron cannot fire even in the case of strong stimulation. Thus, blocking synapses are the ultimate and unconditional version of inhibitory synapses.

## 2.1 The SNN Structure and the General Idea of its Operation

### 2.1.1 The Inference Regime

To simplify explanation how our SNN predicts the approximate time to the next target event, let us assume that weights of all plastic neurons are set somehow to the correct values (the learning process aimed to finding these correct values will be described in the next subsection). In this inference regime, the network receives only sensory spikes carrying information about the current state of the external world (from the lower rectangle labelled "input" on Fig.1). These input spikes come to the learning neurons (denoted by "L") of all columns. However, if the L neuron weights have correct values, at most one L neuron can be active. This is a neuron belonging to the column, which corresponds to the correct time to the next target event (more exactly – this correct time value lays in the interval corresponding to the given column). It is assumed that only one L neuron in this column fires because different L neurons inside one column should react to significantly different world states. The L neuron forces to fire the chain of neurons WTA → ∨ → SECREW interconnected by the strong excitatory links (the fat blue arrows). The intermediate neurons ∨ and SECREW are involved in the learning process – see below. The active ∨ neuron suppresses the SECREW neurons of all columns corresponding to longer predicted times (activity of the leftmost column on Fig.1 signify the shortest expected time to the target event) – see the black arrows going from the ∨ neurons on Fig. 1. Thus, if several columns predict appearance of the target event, the shortest predicted time is selected.

Although the lateral blocking connections in columns go from WTA neurons, they lead to competition between L neurons. However, this competition makes sense only if the competing neurons are significantly different. In the case of "all-to-all" input connections the difference between the L neurons is realized by random initialization of their synaptic weights.

Columns may contain different numbers of neurons. The conditions immediately preceding target events may be more specific and, therefore, would require less neurons to recognize them.

### 2.1.2 The Learning Regime

Now, let us consider how this SNN learns. In this section, we discuss the general idea of the learning mechanism, while formal description of the underlying synaptic plasticity model will be given in the next section.

In the learning process, the spike signals indicating target events are used. At the beginning of the learning process, the weights of all the plastic synapses (connecting the input nodes with the L neurons) are small and, therefore, the L neurons cannot fire. The first target spike comes. It forces the GATE neurons of the leftmost column to fire. Their spikes come to the special plasticity-inducing synapses of the L neurons. When a neuron obtains a spike via its plasticity-inducing synapse, all its plastic synapses which obtained a spike not later than a certain time ago are potentiated by the same constant value. Henceforth, the plasticity-inducing synapses will be referred to as *dopamine* synapses because of their role resembling the rewarding role of biological dopamine synapses. Dopamine connections are denoted by the dashed lines on Fig. 1. Due to this mechanism, after some number of target events, some L neurons in the leftmost column themselves begin to react by firing to the specific combinations of input spikes signaling a soon target event occurrence.

When some L neuron fires, it forces to fire the WTA neuron connected with it (from Fig. 1, it is seen that columns consist of the neuron triplets L – WTA – GATE). Each WTA neuron is connected with the other WTA neurons in the same column by the lateral blocking links. If it fires, it suppresses all other WTA neurons in the same column for some time preventing their firing. Besides that, it suppresses all GATE neurons in its column except the GATE neuron in its own triplet. Thus, this structure behaves in the "winner-takes-all" manner where the winner is the first firing neuron. If the winner, L neuron, fired correctly, after some time a target spike comes. It forces to fire the GATE neuron in the winner's triplet. The GATE neuron rewards the synapses, which contribute to correct firing. The other L neurons in this column are not rewarded because their GATE neurons are blocked and do not pass the reward spike. Moreover, our synaptic plasticity model includes anti-Hebbian component – the synapses contributing to firing are depressed. If the winning L neuron fired correctly then this depression is compensated by subsequent reward, but all other neurons, which fired after the winner remain unrewarded with their contributing synapses depressed. It is right because the different L neurons should recognize the different world states preceding a target event. Due to this mechanism, it is probable that next time other L neurons will not react to the world states, which cause the winner firing.

As it was said in Section 2.1.1, L neuron firing eventually causes SECREW neuron firing which is interpreted as prediction that a target event will happen soon. But in the learning phase, the SECREW neuron firing serves as a reward signal for the right neighboring column. When the leftmost column is trained to predict soon appearance of a target event accurately, spike of its SECREW neuron can play the same role for the second column as the external target spike played for the first, leftmost, column. The learning in the whole SNN goes incrementally – column by column. After some column gets trained to recognize occurrence of a target event at some time distance, the next columns begins to learn to recognize it at longer distance.

The inter-column interaction has one more aspect. If some column predicts a target event, it suppresses all columns corresponding to more distant predictions by blocking connections shown on the upper part of Fig. 1 as right-bound black arrows.

At least, one more necessary feature of the SNN described should be mentioned. If the input connections of the L neurons are fast then the input spikes emitted shortly after a target event may reach an L neuron simultaneously or before the dopamine spike caused by this event. Since these spikes may be not related to the target event, the wrong synapses will be potentiated in this case. To exclude this possibility, we extend the synaptic delays of the input connections to 3 msec (all other connections have 1 msec delay).

## 2.2 Synaptic Plasticity Model in Detail

In the SNN depicted on Fig.1, only the L neurons are plastic. Excitatory synapses of all other neurons have fixed weight values. These values are very high so that an incoming excitatory spike forces the neuron to fire unless it is in the inactive state as it was described just before section 2.1.

Similar to our previous research works [17, 18], the synaptic plasticity rules are additive and applied to a variable termed "synaptic resource," denoted as $W$, rather than directly to the synaptic weight, denoted as $w$. There is functional dependence between $W$ and $w$ expressed by the formula:

$$w = w_{\min} + \frac{(w_{\max} - w_{\min})\max(W, 0)}{w_{\max} - w_{\min} + \max(W, 0)} \tag{3}$$

where $w_{min}$ and $w_{max}$ are constants. It is obvious that $w$ values lay inside the range [$w_{min}$, $w_{max}$) - while $W$ can take any value from -∞ to +∞. In this study, $w_{min} < 0$, while $w_{max} > 0$ so that synaptic plasticity can make excitatory synapse inhibitory and vice versa.

The synaptic plasticity model comprises two distinct and independent components, which are considered in subsections 2.2.1 and 2.2.2.

### 2.2.1 Anti-Hebbian Plasticity

The standard STDP (spike timing dependent plasticity) model [9] states that spikes coming short time before postsynaptic spike emission potentiate the synapses receiving them. This concept aligns with Donald Hebb's principle, which asserts that synaptic plasticity should reflect causal relationships between neuron firings; so that the synapses responsible for inducing neuron firing should be strengthened. Plenty of neurophysiological observations have proved this principle. However, in-depth investigations into plasticity within biological neurons have revealed multiple instances of entirely distinct synaptic plasticity models existing in nature [19, 20]. Furthermore, examples of plasticity rules acting in the direction opposite to Hebbian principle (anti-Hebbian plasticity) have been observed in different organisms [21]. It makes us conclude that different kinds of synaptic plasticity are suitable for the solution of different problems. Besides that, the standard STDP model becomes senseless and self-contradictory in the case (which is quite common in the biological brain) when presynaptic and postsynaptic spikes are not stand alone in time but form tight sequences (spike trains). In this case, it is senseless to say that presynaptic spike comes before or after postsynaptic spike because there are many postsynaptic spikes in the close neighborhood before and after the given presynaptic spike.

For these reasons, we have designed our own variant of the anti-Hebbian plasticity model and applied it to the problem solved.

As it was said, weight modifications in the standard STDP are bound to single pre- and post-synaptic spikes. However, in the presence of spike trains, these rules lose their applicability. Consequently, in our model, the synaptic plasticity acts are bound to postsynaptic spike trains instead of single spikes. We refer to these spike trains as "tight spike sequences" (TSS). Specifically, taking the constant $ISI_{max}$ as a measure of "tightness" of TSS, we define TSS as a sequence of spikes adhering to the following criteria:

1. There were no spikes during time $ISI_{max}$ before the first spike in TSS;
2. Interspike intervals for all neighboring spikes in TSS are not greater than $ISI_{max}$;
3. There are no spikes during time $ISI_{max}$ after the last spike in TSS.

Our anti-Hebbian plasticity model adheres to the following rules:

1. Resource of any synapse can change at most once during a single TSS. Here and below, TSS refers to postsynaptic spike train.
2. Resources of only those synapses are changed which receive at least one spike during TSS or short time $T_H$ before the very first spike in the TSS. Goal of this rule is to weaken all the synapses which contributed to postsynaptic spikes in the given TSS. Therefore, the synapses having obtained spikes shortly before the TSS onset should be also depressed. Effect of one spike to membrane potential decays with the time constant $\tau$, hence $T_H$ should be few times greater than $\tau$ – we selected $T_H = 3\tau$.
3. All synaptic resources are changed (decreased) by the same value $d_H$ independently of exact timing of presynaptic spikes. In this study, $ISI_{max} = L$.

### 2.2.2 Dopamine Plasticity

Every L neuron has a plasticity-inducing (dopamine) synapse connecting the L neuron with the GATE neuron in the same triplet. When a neuron obtains a spike via the dopamine synapse, synaptic resources of its plastic synapses having received at least one presynaptic spike during the time interval of the length $T_P$ in the past are modified (increased) by the same value $d_D$.

In our case, the role of dopamine plasticity is to potentiate all synapses, which could help the neuron to fire in the correct time (during the time L before the dopamine reward). Respectively, setting $T_P = L + 3\tau$ seems to be a reasonable choice.

### 2.2.3 Neuron Stability

In our model, the synaptic plasticity values $d_H$ and $d_D$ are not constant. Instead, they are dynamic and vary in the process of learning. Initially, during the early stages of learning, these values need to be sufficiently large. However, for a trained neuron that consistently makes accurate predictions, they should approach zero. This adaptation is crucial to prevent further modifications to the neuron's synaptic weights, which could potentially disrupt its established trained state. To implement this adaptive behavior, we introduce an additional component of the neuron's state, denoted as "stability."

The synaptic plasticity values decrease exponentially to zero as the stability value grows, in accordance with the following expressions:

$$d_H = \overline{d_H}\min(2^{-s}, 1), \quad d_D = \overline{d_D}\min(2^{-s}, 1). \tag{4}$$

Here, $\overline{d_H}$ and $\overline{d_D}$ are constants of the neuron model. In our SNN, anti-Hebbian and dopamine plasticity should be balanced. Indeed, if a neuron fires in the correct time then its synaptic weights should not be modified – otherwise its trained state could be destroyed. Thus, the anti-Hebbian depression of its synapses should be exactly compensated by the subsequent dopamine potentiation. In order to satisfy this requirement, we set $\overline{d_H} = \overline{d_D}$.

The neuron stability value changes in the two situations:

1. It is decreased by the constant $d_s$ every TSS.
2. It is adjusted by the value $d_s \max\left(2 - \frac{|t_{TSS} - ISI_{max}|}{ISI_{max}}, -1\right)$ whenever a presynaptic dopamine spike is received. Here $t_{TSS}$ is the time interval between the most recent TSS onset and the dopamine spike.

We see that if TSS began exactly the time $ISI_{max}$ (= L) ago before dopamine spike then the neuron stability increase is maximum and is equal to $d_s$ – if to take into account its decrease by $d_s$ in accordance with rule 1. This corresponds to the most accurate prediction of the target event (or the spike from left column's SECREW neuron) and serves as evidence that the neuron is already trained. Conversely, if a dopamine spike occurs when the neuron has remained inactive for a long period, it suggests inadequate training, and as a result, neuron's stability is decreased by $d_s$ to facilitate further learning.

### 2.2.4 Constant Total Synaptic Resource

In order to introduce the competition between synapses and to avoid possible unlimited neuron excitation, we added one more component to our model of synaptic plasticity – constancy of neuron's total synaptic resource. Whenever some synapses are depressed or potentiated due to the above mentioned plasticity rules all the other synapse are changed in the opposite direction by the constant value equal for all synapses such that to preserve to total synaptic resource of the neuron. Effect of this rule can be controlled introducing imaginary unconnected synapses whose only role is to be a reservoir for the excessive (or additional) resource. The competitive effect is maximum when there are no such silent synapses and it vanishes with their number approaching infinity.

## 2.3 The Test Task – to Predict Obtaining Reward in the Ping-Pong ATARI Game

The described technique holds great promise in the realm of reinforcement learning (RL). In many RL tasks, reward signals come rarely, and, therefore, binding synaptic plasticity acts exclusively to them would lead to slow learning. In order to overcome problems with insufficient evaluation signal frequency, the mechanism of intermediate goals should be utilized. This mechanism should be based on a predictive model, which evaluates the current world state from the view point of its proximity to reward – if the world state gets closer to a desired state marked by reward signal as a result of agent's action, it means that this action was correct and should be evaluated by an artificial internal reward signal. Wrong actions can be determined in the similar way. The proposed future event prediction mechanism can serve as a natural basis for generation of these intermediate evaluation signals. Keeping this in mind, we selected the first task to test this mechanism from the RL domain.

For this purpose, we chose one of the RL tasks drawn from the ATARI games benchmark set [22] often used for RL implementation testing, namely, the ping-pong game. In this game, a ball traverses within a square area, rebounding off its walls. The area has only three walls. Instead of the left wall, the racket moves in the vertical direction on the left border of this square area. The racket is controlled by the agent which can move it up and down. When the ball hits the racket it bounces back and the agent obtains a reward signal. If the ball crosses the left border without hitting the racket the agent gets punishment and the ball is returned to a random point of the middle vertical line of the area, gets random movement direction and speed and the game continues. Using the reward/punishment signals received, the agent should understand that its aim is to reflect the ball and learn how to do it.

In our example, the network's task is related to the former problem – to "understand" which world states are closer to obtaining reward (= make it more probable to reflect the ball) and which are farther.

Let us describe the input information coming to plastic synapses of the L neurons. This information includes the current positions of the ball and the racket and the ball velocity. While the ultimate formulation of this problem would involve primary raster information (i.e., the screen image), computer vision is not in our primary focus in this study whose main goal is future event prediction. Consequently, we assume that preceding network layers have already processed the primary raster data and converted it into the spike-based description of the world state, which forms the basis of the L neuron input.

The input nodes that are sources of spikes sent to the learning neurons are subdivided into the following sections:

1. The ball X coordinate Consists of <u>30 nodes</u> capturing the ball's horizontal position<u>.</u> The horizontal dimension is broken to 30 bins. When the ball is in the bin *i*, the *i*-th node emits spikes with frequency 300 Hz. To establish spatial and temporal scales we assume that the size of the square area is 10×10 cm (so that the boundary coordinates are ±5 cm) and the discrete emulation time step is 1 msec.
2. The ball Y coordinate. Consists of <u>30 nodes</u> capturing the ball's vertical position<u>.</u> Similar to X but for the vertical axis.
3. The ball velocity X component. Consists of <u>9 nodes</u> capturing the ball's horizontal velocity<u>.</u> When the ball is reset in the middle of the square area, its velocity is set to the random value from the range [10, 33.3] cm/sec. Its original movement direction is also random but it is selected so that its X component would not be less than 10 cm/sec. The whole range of possible ball velocity X component values is broken to 9 bins such that the probabilities to find the ball at a random time moment in each of these bins are approximately equal. While the ball X velocity is in some bin, the respective input node emits spikes with 300 Hz frequency.

4. The ball velocity Y component. Consists of 9 nodes capturing the ball's vertical velocity. The same logic as for the X velocity component.
5. The racket Y coordinate. Consists of 30 nodes capturing the racket's vertical position. Similar to the ball Y coordinate. The racket size is 1.8 cm so that the racket takes slightly more than 5 vertical bins.
6. The relative position of the ball and the racket in the close zone. Consists of 25 nodes capturing the ball's positions close to the racket. The square visual field of size 3×3 cm moves together with the racket so that the racket center is always at the center of the left border of this visual field. The visual field is broken to 5×5 square zones. When the ball is in some zone, the respective input node fires with frequency 300 Hz.

In total, there are 133 input nodes transmitting their spikes to the learning neuron. The SNN's objective was to discern the world states after which a reward signal would be obtained in the three proximity intervals [0 – 100], [100 – 200] and [200 – 300] milliseconds. Thus, in this task, $N = 3$, $L = 100$ msec.

## 2.4 Selection of the Network Parameters Using the Genetic Algorithm

Similar to traditional neural network, SNNs includes hyperparameters, which require tuning for the task set selected. In our model, the following network parameters should be optimized:

- The number of L neurons in a column – $n_0$.
- The time constant $\tau$ of the L neurons.
- The silent synapse count (see section 2.2.4) for the L neurons $N_s$.
- The maximum synaptic resource change $\overline{d_H}$. This parameter controls learning speed. Its low values make learning slow, high values may make it unstable.
- The minimum synaptic weight value $w_{min}$. It is negative.
- The maximum synaptic weight value $w_{max}$.
- The ratio $r_s = d_s / \overline{d_H}$.

Since the general theoretical principles for setting these parameters are insufficiently clear, we decided to find their optimum values using the genetic algorithm and the very wide search ranges. They are shown (together with the best values found) in Table 1.

For setting these random values, the log-uniform distribution was used (except the last one in Table 1). The genetic algorithm parameters were following: the population size was 300; the elitism level was 0.1; the mutation probability per chromosome was 0.5. The optimization criterion was $R^2$ (1). It was measured for the last 600 sec of 2000 sec record of ping-pong game where the racket moved chaotically. The total number of rewards was 951. In order to make estimation of $R^2$ more reliable, the mean value of $R^2$ in 3 experiments with the same network hyperparameters but different initial synaptic resource values were taken as a criterion value. The genetic algorithm terminated when 3 successive generations showed no $R^2$ increase.

**Table 1. The SNN parameters optimized.**

| Parameter optimized | Value range | Optimum value |
|---|---|---|
| The number of L neurons in one column $n_0$ | 1-30 | 1 |
| The time constant of the L neurons $\tau$ | 1-30 msec | 1 msec |
| The silent synapse count for the L neurons $N_s$ | 1-300 | 118 |
| The maximum synaptic resource change $\overline{d_H}$. | 0.03 – 1 | 0.049 |
| The minimum synaptic weight value $w_{min}$ | -0.003 – -1 | -0.019 |
| The maximum synaptic weight value $w_{max}$ | 0.03 – 1 | 0.45 |
| Ratio $r_s$ | Normal distribution with the center in 0 and the standard deviation 3. Negative value means that the stabilization mechanism is not used. | 0.487 |

# 3 Results and Discussion

The best result was reached in the 16[th] generation of genetic algorithm. The $R^2$ value reached was 0.707.

The optimum hyperparameter values are summarized in Table 1. Surprisingly, the optimum value of $\tau$ was found to be equal to 1 msec. Since the emulation time step in our experiments was 1 msec, it means that the membrane potential of the L neurons is zeroed before every simulation iteration. Therefore, even the simplest binary neurons can be used as L neurons. The second interesting finding is the fact that the best network is very small. It contains only one learning neuron per column. The solved problem is not very complex but it is not trivial, either. In our opinion, the results obtained by such a small network in this task is an evidence of high potential of the SNN model proposed in this paper.

Considering the inherent fuzziness in the relationship between the current world state and soon obtaining reward due to the discrete world description and chaotic racket movement, the $R^2$ value demonstrated by the network appears quite high. To objectively evaluate the performance, we decided to compare our SNN with traditional machine learning methods on this task. To ensure a fair comparison, we selected two machine learning algorithms of completely different nature: decision tree and convolutional neural network. All the algorithms were trained on the same binary signal data from the input nodes, with each emulation step serving as a learning example. The target value was $P(t)$ from (2). The machine learning algorithms were applied on the same data as our network (the first 1 400 000 steps) and created the predictive models for $P(t)$. These models were tested on the last 600 000 simulation steps. The decision tree algorithm used the information gain split criterion. The network included 2 convolutional ReLU layers.

The maximum $R^2$ value obtained by decision tree is 0.57, convolutional network gives $R^2$ equal to 0.552. We can conclude that the specific structure of our SNN allows it to outperform the traditional machine learning methods in this task. The machine learning algorithms used only the current input information for prediction – but it was true for the learning L neurons, as well. Since their characteristic time was 1 msec, they also cannot "remember" event the recent past – their firing is exclusively determined by the spikes coming to them at the current emulation step.

Let's examine the learning process and its outcomes. The dynamics of the L neuron activities are represented on Fig. 2. We see that the columns are switched on one after one – from left to right (Fig. 1). At first, only the primary reward signals come to the L neurons. After some period of learning, the weights of the leftmost L neuron become sufficiently great to make its firing possible. It causes generation of internal reward signals and the process propagates to right neighboring columns.

The synaptic weight dynamics show the similar picture (Fig. 3). On the beginning stages of learning, weights of the L neuron obtaining external reward signal change. The other L neurons start to learn as sources of internal reward signals gradually become active. After a certain learning period, the L neuron weights "freeze" because of the weight stabilization mechanism whose effect is shown on Fig. 4. It is seen that stability of all L neurons increases almost monotonously.

The learning results are presented on Fig. 5 depicting values of synaptic resources of the L neurons at the 2000[th] second. The leftmost plots correspond to 30 input nodes coding the ball X coordinate. The vertical axis of all plots except the rightmost ones displays synaptic resource value. The second plot column corresponds to 30 input nodes coding the Y coordinate of the ball (the blue line) and the racket (the orange line). The next two plot columns represent 9+9 input nodes coding the horizontal and vertical components of the ball velocity. The rightmost plot column shows the color-coded values of synaptic resources of the 25 input nodes indicating the location of the ball within a 5x5 grid that moves with the racket. The distribution of synaptic resource values in these plots appears reasonable and in line with expectations.

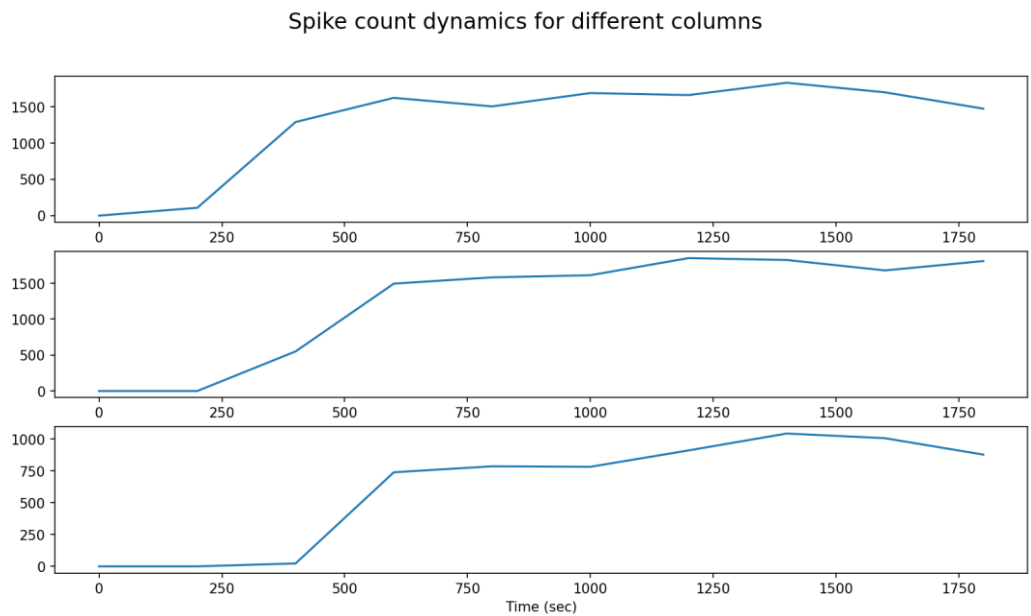

Fig. 2. Dynamics of firing frequency of the L neurons from different columns. The upper plot corresponds to the leftmost column on Fig. 2.

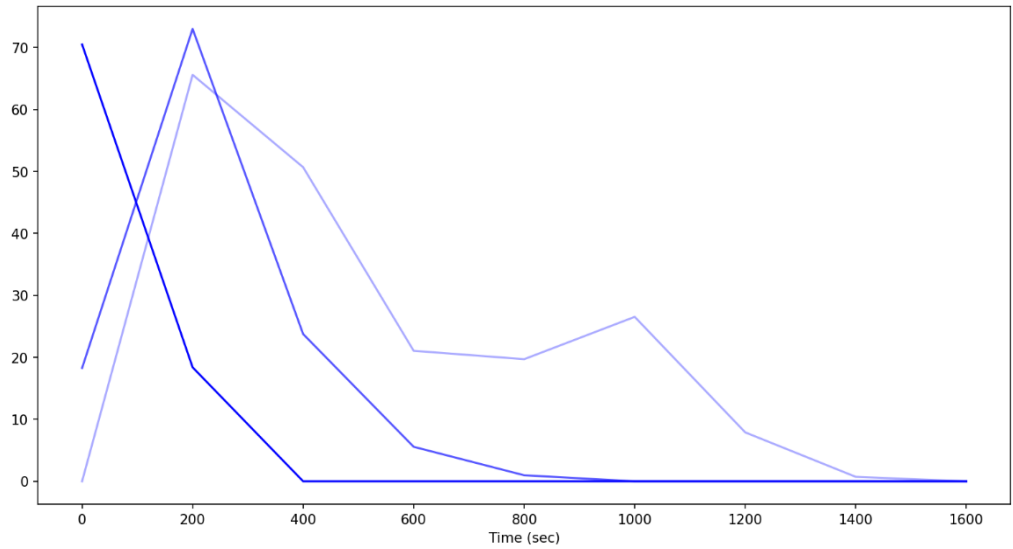

Fig. 3. The sum of the synaptic weight change absolute values for the L neurons from different columns. The darkest line corresponds to the leftmost column on Fig. 2.

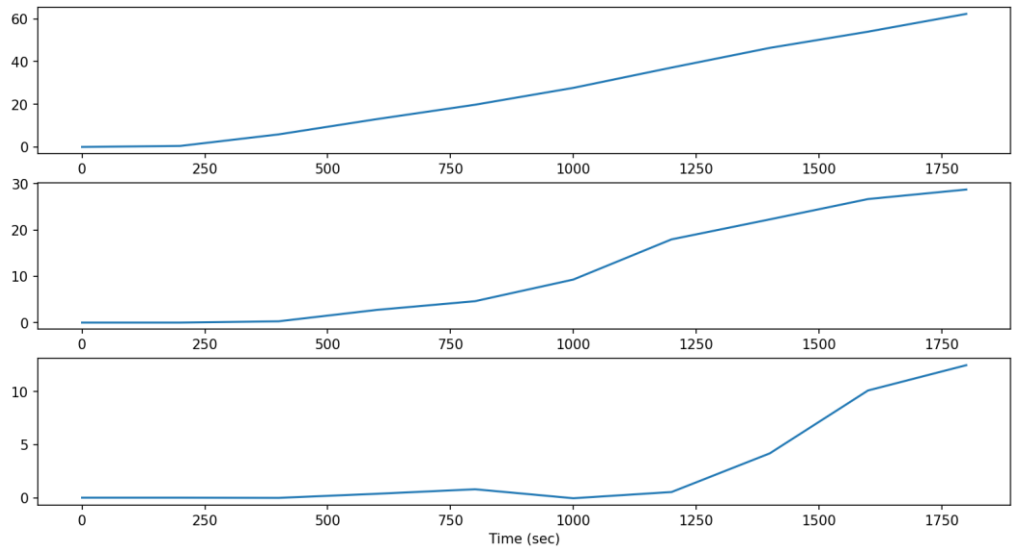

Fig. 4. Dynamics of stability of the L neurons from different columns. The upper plot corresponds to the leftmost column on Fig. 2.

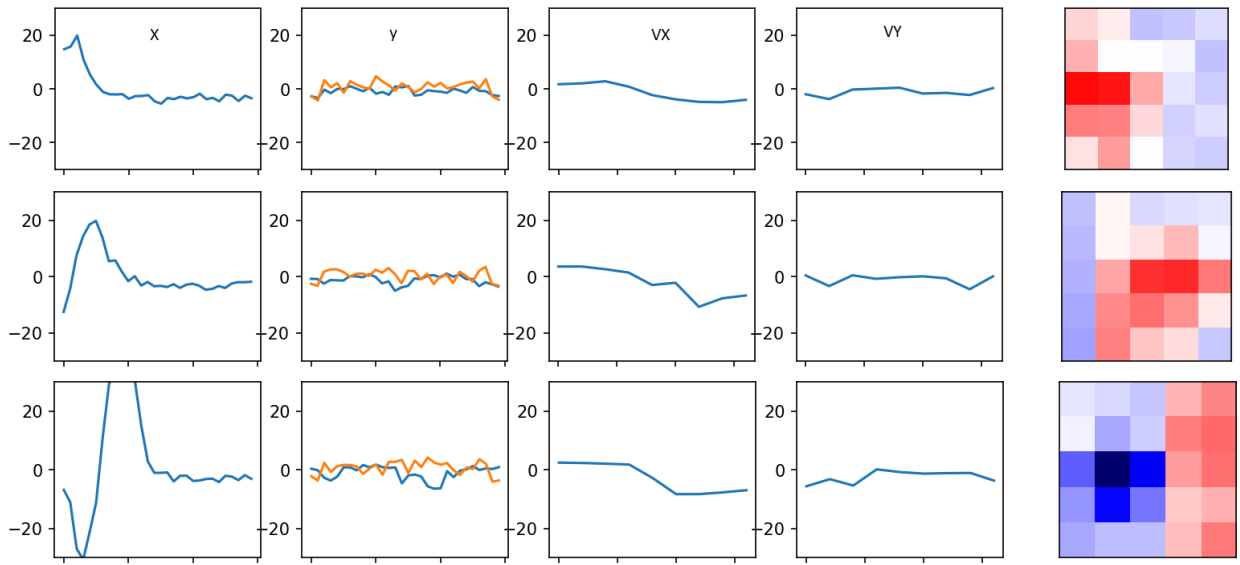

Fig. 5. Synaptic resources of the trained L neurons. The upper row corresponds to the leftmost column on Fig. 2.

These results enable us to assert that our SNN is capable of assessing the probable time to the next target event.

## 4    Conclusion.

The ability to determine the approximate time to appearance of events of interest is a fundamental feature for any learning system interacting with dynamic processes of the real world. In this research, we have demonstrated the realization of this critical function in the SNN with a specially designed "columnar" structure and a novel combination of anti-Hebbian and dopamine plasticity mechanisms. We tested our model on a simplified RL problem, namely, the ATARI ping-pong computer game, where the network should learn to estimate when the next reward signal may come. Our findings, including the comparison of the prediction accuracy reached by our network and by two different traditional machine learning methods in this task, enable us to declare high efficiency of the proposed models of SNN and synaptic plasticity in this class of learning tasks.

In the forthcoming research, we aim to include the neuronal structures considered in the present work into the SNN solving the reinforcement learning problems. We believe that this research represents a significant step toward the purely spiking implementation of model-based reinforcement learning.

## References.